\documentclass[runningheads]{llncs}
\usepackage{graphicx}

\usepackage{tikz}
\usepackage{comment}
\usepackage{amsmath,amssymb} 
\usepackage{color}
\usepackage{mathtools}
\usepackage{multirow}
\usepackage{arydshln}
\usepackage{bbding}
\usepackage{pifont}
\usepackage{wasysym}

\usepackage{ulem}

\usepackage[accsupp]{axessibility}  

\begin{document}
\pagestyle{headings}
\mainmatter
\def\ECCVSubNumber{1298}  

\title{CLIP2TV: Align, Match and Distill for Video-Text Retrieval}

\titlerunning{CLIP2TV: Align, Match and Distill for Video-Text Retrieval}
%
\author{Zijian Gao\inst{1} \and
Jingyu Liu\inst{1} \and
Weiqi Sun\inst{1} \and \\
Sheng Chen\inst{1} \and
Dedan Chang\inst{1} \and
Lili Zhao\inst{1}}
\authorrunning{Zi. Gao et al.}
%
\institute{OVBU, PCG, Tencent\\
\email{gzjbupt2016@bupt.edu.cn, sunweiqi@buaa.edu.cn, \\ \{messijyliu, carlschen, dedanchang, lilillzhao\}@tencent.com}}
\maketitle

\begin{abstract}
Modern video-text retrieval frameworks basically consist of three parts: video encoder, text encoder and the similarity head.
With the success of both visual and textual representation learning, transformer-based encoders and fusion methods have also been adopted in the field of video-text retrieval.
In this paper, We propose a new CLIP-based framework called CLIP2TV, which consists of a video-text alignment module and a video-text matching module. The two modules are trained end-to-end in a coordinated manner, and boost the performance to each other. Moreover, to address the impairment brought by data noise, especially false negatives introduced by vague description in some datasets, we propose similarity distillation to alleviate the problem.
Extensive experimental results on various datasets validate the effectiveness of the proposed methods. Finally, on common datasets of various length of video clips, CLIP2TV achieves better or competitive results towards previous SOTA methods.
\keywords{Video-Text Retrieval, Multi-Modal Learning}
\end{abstract}
\section{Introduction}

Multi-modal retrieval is a fundamental task in both research fields of multi-modal learning and industrial applications. Specifically, visual-text retrieval is attracting more attention to meet the needs of increasing tons of uploaded texts, images and videos. This work focuses on the task of video-text retrieval. Recently, both the research community \cite{xu2016msrvtt,chen2011collecting,wang2019vatex,krishna2017dense,zhang2018cross,anne2017localizing,liu2019use,bain2021frozen} and the industry are putting more efforts into it. Given the textual input, the system is required to retrieve corresponding videos, which complements traditional search engines relying on pure textual labels. In reverse, given the video input, the system could also output textual descriptions, which is useful for videos without captions.

A modern video-retrieval system typically consists of three parts: text encoder, video encoder, and similarity head. With the success of transformers \cite{vaswani2017attention} in natural language processing, visual representation learning \cite{dosovitskiy2020image}, and multi-modal learning \cite{chen2020uniter}, the three parts have all been benefited from transformer mechanisms. For text encoder, from BERT-like models pretrained on pure texts, to CLIP-like models pretrained on texts and images, downstream tasks can exploit them either via fine-tuning or prompt learning. For video encoder, transformer-based methods are also showing great potentials. For instance, ViT \cite{dosovitskiy2020image} and video Swin-Transformer \cite{liu2021video-swin} are used in action recognition and other related tasks. For similarity head, which is also interpreted as multi-modal fusion, can be implemented as multi-modal transformers. Heavy models like ViLBERT \cite{lu2019vilbert} and UNITER \cite{chen2020uniter} fuse texts and images as input tokens, with self and mutual attention learning via heavy layers of attention networks. To fully realize the potential of transformer-based backbones, we adopt text and image encoders from CLIP \cite{radford2021clip} as text and video encoders in our proposed CLIP2TV.

\begin{figure}[t]
    \centering
    \includegraphics[width=0.98\linewidth]{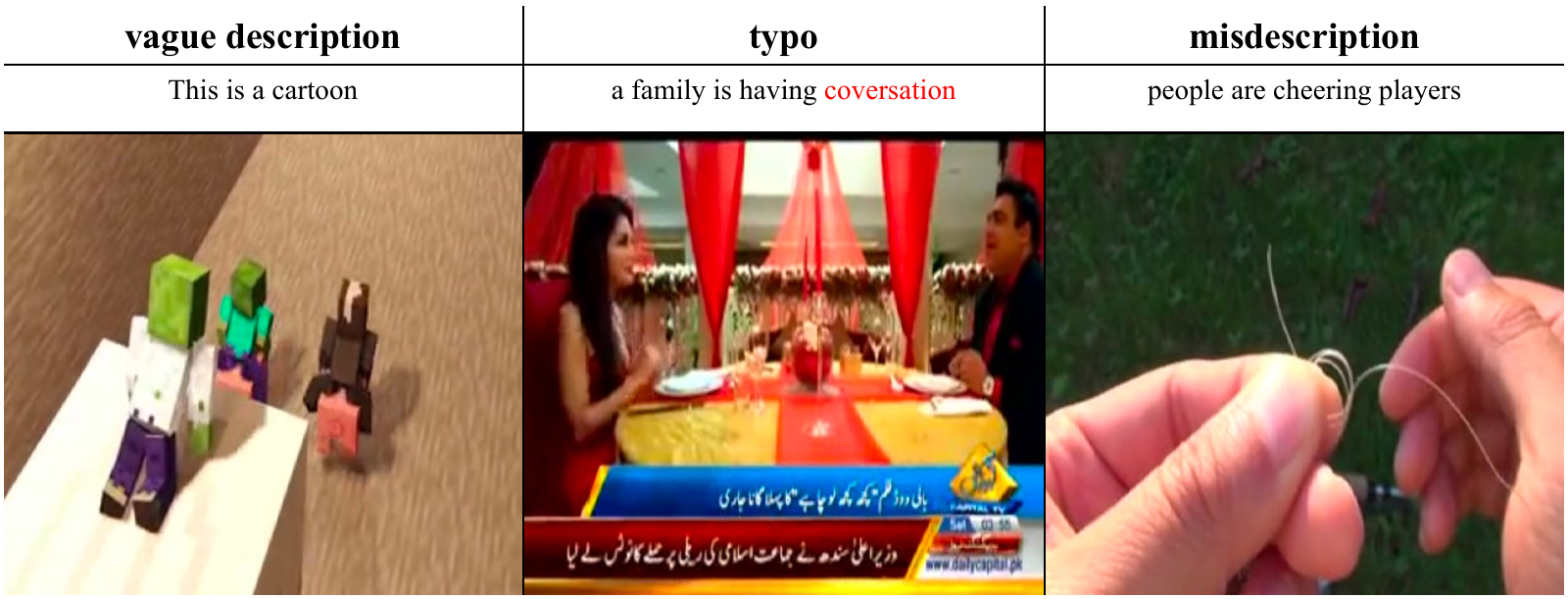}
    \caption{\small{Three types of noise exist in descriptions. \textbf{Vague description}, or referred to low-quality description, can describe multiple distinct videos. \textbf{Typo} will result in the loss of some key information. \textbf{Misdescription} is a false description of some video contents or an error caption that is completely irrelevant to the video.}}
    \label{fig: noise}
\end{figure}

Basically, our work follows the line of recent CLIP-based models \cite{luo2021clip4clip}, \cite{fang2021clip2video}. A similarity head is added on two streams of visual and textual encoders. The outputs of these two modules are projected and aligned in the common multi-modal space, to formulate the video-text alignment (vta) module. Besides it, we are different from \cite{luo2021clip4clip} in two respects:

(i) For similarity head, we propose a video-text matching (vtm) module initialized from multiple layers of transformers. The vtm module takes as inputs online positive and negative pairs of embeddings from text and video encoders, and formulates a matching loss like previous multi-modal transformers \cite{lu2019vilbert}, \cite{chen2020uniter}. In \cite{luo2021clip4clip}, a transformer head is also added on two streams, but shows worse results towards simpler heads. We claim that alignment in the first place is critical to make vtm work. On the other hand, by back propagating the gradients in an end-to-end manner, the vtm module can improve the accuracy of vta. This two modules work in a coordinated manner and boost performance to each other. Similar structure in image-text retrieval is \cite{li2021align}, which also uses an image-text alignment module to input embeddings to the image-text matching module. Instead of \cite{li2021align} using matching score of the similarity head of vtm as the final retrieval result, we still use nearest neighbors in the common space from vta as the retrieval results. Therefore CLIP2TV is efficient for inference.

(ii) In the training process, we observe that vtm is sensitive to noisy data thus oscillates in terms of validation accuracy. We attribute the cause of it to the data noise. There are mainly three types of noise in video-text retrieval datasets: vague description, typo and misdescription, as showed in Fig.\ref{fig: noise}. Among them, vague descriptions are the most frequent. For instance, in MSR-VTT \cite{xu2016msrvtt}, some descriptions are too general like ``This is a cartoon'' for a cartoon video. Also in DiDeMo \cite{anne2017localizing}, texts like ``we see only the sky'' make the target video hardly stand out. This will impair the contrastive training process where negative samples might be semantically close to the anchor sample. 
Typo and misdescription also exist, which will result in inaccurate description of videos and impair the relation between the paired text and video.
To alleviate this problem, we aim to find the metric that reflects more accurately the similarity between two modalities. Therefore we propose similarity distillation, which distills 3 types of similarities from inter-, intra- or joint space before vtm. Then we use them as additional soft labels to supervise vtm.

In summary, we propose CLIP2TV, a new CLIP-based framework to address video-text retrieval. Our contributions are threefold:

1. The framework is comprised of two modules: video-text alignment (vta) and video-text matching (vtm). They are trained end-to-end in a coordinated manner and benefit each other.

2. With data exploration and extensive experiments, we find that network training suffers impairment brought by data noise, especially videos with vague descriptions. Therefore we propose similarity distillation to alleviate the problem.

3. We conduct comprehensive experiments on multiple datasets, achieving significant improvements on the strong baseline of CLIP4Clip, and finally get better or competitive results to previous SOTA methods on common datasets. We believe this work can bring useful insights and practical expertise to both research community and industry.
\section{Related Work}

\noindent\textbf{Video-Language Modeling. }
VideoBERT \cite{sun2019videobert} firstly extends BERT \cite{vaswani2017attention} to video representation learning and adapts it for modeling visual-linguistic relationship and joint representation learning. The following work of CBT \cite{sun2019cbt} lightens VideoBERT \cite{sun2019videobert} by combining video and text representations after separate modality training. Then HERO \cite{li2020hero} proposes a hierarchical structure for encoding video and subtitle/speech together. While ActBERT \cite{zhu2020actbert} utilizes global action information to bridge the gap between video and language and introduces transformer block to encode global actions, local objects, and text descriptions. Different from canonical approaches, ClipBERT \cite{lei2021less} learns video and text representations by an end-to-end manner, and adopts the sparse sampling strategy and initializes model with pretrained 2D model for efficiency and lower computational burden. Further more, to fully exploit the advantage of transformer, Frozen \cite{bain2021frozen} leverages a dual transformer encoder structure with end-to-end training. Nowadays, models such as CLIP4Clip \cite{luo2021clip4clip} and CLIP2Video \cite{fang2021clip2video} extended from a large-scale image-text pretrained model CLIP \cite{radford2021clip} are continuously proposed.

\noindent\textbf{Image/Video-Text Retrieval. }
Multi-modal retrieval, as one of the most popular tasks in multi-modal learning, attracts lots of attention to produce valuable works \cite{lu2019vilbert,chen2020uniter,radford2021clip,li2021align,lei2020tvr,li2020hero,yang2021taco,liu2021hit,luo2021clip4clip,fang2021clip2video}. ViLBERT \cite{lu2019vilbert} and UNITER \cite{chen2020uniter} use a shared transformer for image-text joint representation learning and retrieval. CLIP \cite{radford2021clip} uses two independent transformer to encode image and text separately, yet ALBEF \cite{li2021align} adopts pyramid-like structure that images and texts are separately encoded in shallow layers and fused in deeper layers for matching. Recently, video-text retrieval task earns much more attention. HiT \cite{liu2021hit} and TACo \cite{yang2021taco} use multi-level feature alignment for better cross-modal learning. Moreover, TVR \cite{lei2020tvr} and HERO \cite{li2020hero} extend video-text retrieval task to video corpus moment retrieval task aiming to retrieval a whole video from a large video corpus and localize the related moment within the video by a query. Meanwhile, inspired by image-text retrieval, many studies \cite{bain2021frozen,luo2021clip4clip,fang2021clip2video} try to transfer knowledge from large-scale image-text pretraining model into video-text retrieval or train models with both images and videos.

\noindent\textbf{Learning with Noise. }
Learning with noise is quite popular in recent years. Many works focus on supervised learning from noise\cite{reed2014training,patrini2017making,li2017learning,han2018co,zhang2018generalized,chuang2022robust,hoffmann2022ranking,huang2021learning,hu2021learning}. Since deep neural networks can easily overfit to noisy labels \cite{zhang2021understanding}, which results in poor generalization.
Several techniques have been developed to enhance the robustness to label noise, including losses that reduce the impact of outliers \cite{zhang2018generalized,wang2019symmetric,ghosh2017robust}, loss correction approaches that model the source of label noise \cite{patrini2017making,reed2014training,hendrycks2018using,song2019selfie,arazo2019unsupervised}, and regularization procedures tailored to lower the impact of noise \cite{zhang2017mixup,pereyra2017regularizing}. We refer readers to \cite{song2020learning} for a detailed survey of prior work on learning with label noise. In this work, we consider that existing video retrieval benchmarks are based on datasets collected for other tasks (e.g. video captioning)~\cite{wang2022multi-query}, which often leads to ambiguity of the captions (i.e. can be matched to many videos).
\cite{wang2022multi-query} proposes multi-query training and inference method to reduce the impact of vague or low quality queries.
\section{Proposed Method}

\begin{figure}[t]
    \centering
    \includegraphics[width=0.95\linewidth]{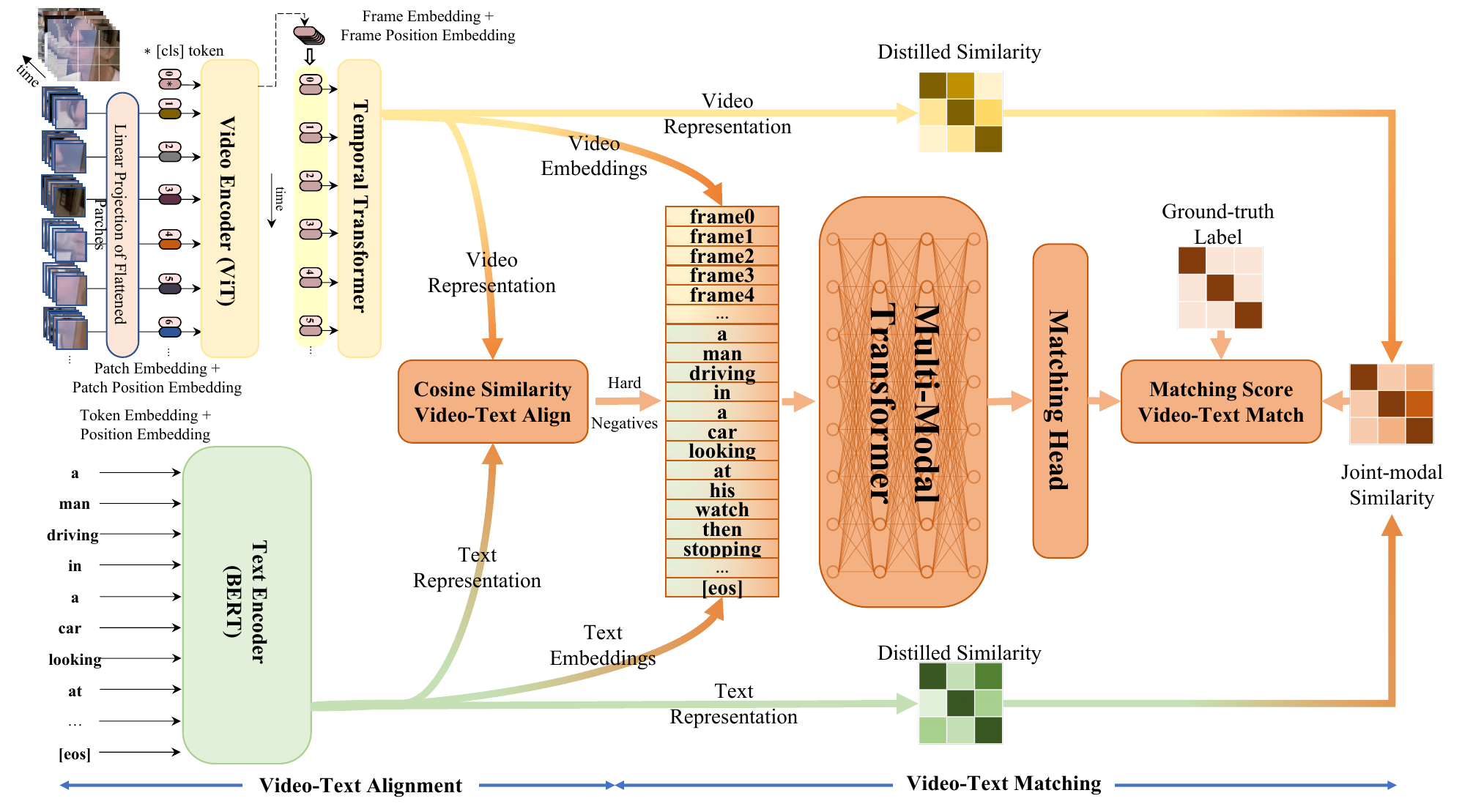}
    \caption{\textbf{Framework of our method.} Vta: Dual transformer encoders encode video and text representations, which are then projected and aligned in the common multi-modal space. Vtm: The following multi-modal transformer and matching head take as input concatenated video-text embeddings and output the matching score. Similarity distillation: distilled similarity from intra- or inter-space serve as soft labels for video-text matching.}
    \label{fig: pipeline}
    \vspace{-5mm}
\end{figure}

Given a set of captions and a set of videos, video-text retrieval aims at seeking a matching function calculating similarities between captions and videos. Recent works like \cite{luo2021clip4clip} have showed the benefits of image-text retrieval pre-training and advantages of end-to-end training. Inspired by these, we adopt CLIP \cite{radford2021clip} as our multi-modal encoder and base our work on \cite{luo2021clip4clip}. Fig.\ref{fig: pipeline} depicts the framework of CLIP2TV. It is comprised of a video-text alignment (vta) module and a video-text matching (vtm) module. In the following, we show how vta and vtm interact with each other in detail, and how similarity distillation address the problem caused by noisy samples. 

\subsection{Framework}

\noindent\textbf{Video Encoder:} Given a video, we first split it into equal-length segments, then randomly extract one frame from each segment. For a video clip $v = \{v_i\}^{N_v}_{i=1}$, $N_v$ is number of frames. A transformer-based Video Encoder, i.e., ViT used in CLIP, encodes it into frame embeddings $\mathbf{V}^{emb} \in \mathbb{R}^{N_v \times D_v}$, where $D_v$ is the dimension of frame embedding. After that, a temporal transformer with frame position embedding and residual connection is applied to frame embeddings to enhance temporal information. Then, a MLP projector projects contextualized frame embeddings into a multi-modal common space. For simplicity,we re-use $\mathbf{V}^{emb} \in \mathbb{R}^{N_v \times D}$ as enhanced temporal embeddings, where $D$ is the dimension. The final video embedding is calculated as $\mathbf{v} = \texttt{mean-pooling}(\mathbf{V}^{emb}) \in \mathbb{R}^D$.

\noindent\textbf{Text Encoder:} The bert-like transformer used in CLIP is adopted as our text encoder. The caption denoted as $t = \{t_i\}^{N_t}_{i=1}$ is fed into encoder to acquire token embeddings $\mathbf{T}^{emb} \in \mathbb{R}^{N_t \times D_t}$, where $N_t$ and $D_t$ are the number and dimension of token embeddings. Same with the procedure of encoding videos, a text projector projects token embeddings into a common sub space with dimension $D$. Finally, the representation of [EOS] token will serve as the embedding of the whole caption $\mathbf{t} \in \mathbb{R}^D$.
 
\noindent\textbf{Contrastive Learning by Alignment (vta):} To align video embeddings $\mathbf{v}$ and caption embeddings $\mathbf{t}$ in the common space, we adopt contrastive loss to maximize similarity between positive pairs and minimize similarity between negative pairs. The cosine similarity is calculated between normalized embeddings from each modality. Given a batch $\mathcal{B}$ with $B$ video-text pairs, cross-entropy loss adds on softmax-normalized similarity gives the InfoNCE loss:
\begin{align}
    p^{v2t}_{vta}(v)&\sim\frac{\exp(s(\mathbf{v},\mathbf{t})/\tau_1)}{\sum_{\mathbf{t} \in \mathcal{B}}\exp(s(\mathbf{v},\mathbf{t})/\tau_1)}, \label{eq:prob}\\
    \mathcal{L}^{v2t}_{vta}&=\mathbb{E}_{v\sim\mathcal{B}}\left[H(\mathbf{p}^{v2t}_{vta}(v), \mathbf{y}^{v2t}_{vta}(v))\right], \label{eq:contrast}
\end{align}
where $s(\cdot, \cdot)$ denotes cosine similarity, $\tau_1$ represents a learnable temperature parameter, $\mathbf{y}^{v2t}_{vta}(\cdot)$ is the ground-truth binary label that positive pairs and negative pairs are $1$ and $0$ respectively, and $H(\cdot, \cdot)$ is cross-entropy formulation. Meanwhile, $\mathcal{L}_{t2v}$ is calculated vice versa. Then, the contrastive loss for alignment is formulated as:
\begin{equation}
    \mathcal{L}_{vta} = \frac{1}{2}(\mathcal{L}^{v2t}_{vta}+\mathcal{L}^{t2v}_{vta})
\end{equation}


\noindent\textbf{Contrastive Learning by Matching (vtm):} To fully exploit the abundant information between two modalities and enhance cross-modal interactions, we utilize a multi-modal fusion and matching strategy to predict whether a video-text pair is positive or negative. Similar with TACo \cite{yang2021taco} and ALBEF \cite{li2021align}, our \textbf{multi-modal encoder} consists of self-attention layers. It takes as input video frame embedding $\mathbf{V}^{emb} \in \mathbb{R}^{N_v\times D}$ and caption token embedding $\mathbf{T}^{emb} \in \mathbb{R}^{N_t\times D}$ and outputs fusion embedding $\mathbf{F}^{emb} \in \mathbb{R}^{(N_v+N_t)\times D}$. We adopt embedding of [EOS] token as fusion feature of video-text pairs, which is $\mathbf{f} \in \mathbb{R}^{D}$. Then, a matching head $\mathit{g}(\cdot)$ composed of \texttt{LN-FC-ReLU-FC} calculates the matching score. The training objective of multi-modal fusion is also InfoNCE loss:
\begin{align}
    p_{vtm}(v,t)&\sim\frac{\exp(\mathit{g}(\mathbf{f}_{(v,t)}))}{\sum_{(v',t')\in\mathcal{P}}\exp(\mathit{g}(\mathbf{f}_{(v',t')}))},\\
    \mathcal{L}_{vtm}&=\mathbb{E}_{(v,t)\in\mathcal{B}}\left[H(\mathbf{p}_{vtm}(v,t),\mathbf{y}_{vtm}(v,t))\right],
\end{align}
where $\mathcal{P}$ denotes the pair set that consists of a ground truth video-text pair and the negative sample pairs related to them. $p_{vtm}(\cdot, \cdot)$ represents the probability whether a pair in the pair set is positive. And $\mathbf{y}_{vtm}(\cdot,\cdot)$ is the ground-truth one-hot label of the pair set, while $H(\cdot, \cdot)$ is cross-entropy formulation.

In order to improve efficiency and reduce computational cost, we only select in-batch hard negative samples for multi-modal fusion. Concretely, for each video, we choose top-$K$ negative text samples during real-time training according to eq.\eqref{eq:prob} and vice versa. Finally, $B\times(2K+1)$ pairs are fed into multi-modal fusion.

\subsection{Similarity Distillation on vtm}
As mentioned before, in the training process, we observe that vtm suffers severer oscillation than vta, and ends up with lower accuracy. We attribute this to two reasons: 1. Data noise including vague description, typo and misdescription. Among them vague description is the most frequent. Since most datasets are originally designed for video captioning, which are not particularly labeled to be distinct. This problem is also mentioned in \cite{wang2022multi-query}. 2. The inherent nature of multi-modal transformer in vtm. For instance, anchor textual embedding and positive/negative video embedding are fused in an earlier stage, thus all the final outputs contain the information of the anchor embedding, which results in more difficulty in separation.

To seek for a more accurate metric on the similarity between positive-positive pairs and positive-negative pairs, we refer to the output embeddings from vta. There are three sub-spaces in vta: textual space, video space and the aligned multi-modal space. Each space has its own metric on the similarity of a pair of instances. This raises curiosity to us: can we refer to the similarity in them and which will be the best.
Without loss of generality, we will try three variants of strategies.

1.\textit{\textbf{Inter-modal}}: 
It is natural to adopt inter-modal similarity from vta as the soft label to guide video-text matching as:

\begin{equation}
    \mathbf{y}_{inter}^{t2v} = \mbox{softmax}\left(s(\mathbf{t},\mathbf{v'})/\tau_2\right)
    \quad \textrm{and} \quad
    \mathbf{y}_{inter}^{v2t} = \mbox{softmax}\left(s(\mathbf{v},\mathbf{t'})/\tau_2\right)
\label{eq:y_inter}
\end{equation}
where $\mathbf{y}_{inter}$ denotes inter soft-label. $\tau_2$ is temperature to control distillation. $\mathbf{v}$ and $\mathbf{t}$ denote video embedding and text embedding, respectively.

2.\textit{\textbf{Intra-modal}}: We utilize video-video similarity in video space on text-to-video matching, and text-text similarity in textual space on video-to-text matching. Formally,

\begin{equation}
    \mathbf{y}_{intra}^{t2v} = \mbox{softmax}\left(s(\mathbf{v},\mathbf{v'})/\tau_2\right)
    \quad \textrm{and} \quad
    \mathbf{y}_{intra}^{v2t} = \mbox{softmax}\left(s(\mathbf{t},\mathbf{t'})/\tau_2\right)
\label{eq:y_intra}
\end{equation}
where $\mathbf{y}_{intra}$ denotes intra soft-label.  

3.\textit{\textbf{Joint-modal}}: Joint soft-label utilize intra-modal similarity distilled knowledge from video-modal space and text-modal space jointly to supervise vtm module. Concretely, we compute average intra-modal similarity of both modalities to obtain the joint soft-label:

\begin{equation}
    \mathbf{y}_{joint}^{t2v,v2t} = \mbox{softmax}\left(s(\mathbf{t},\mathbf{t'}) + s(\mathbf{v},\mathbf{v'})/2\tau_2\right)
\label{eq:y_joint}
\end{equation}
where $\mathbf{y}_{joint}$ denotes joint soft-label.

Then we compute soft InfoNCE loss for multi-modal encoder:
\begin{equation}
    \mathcal{L}_{soft-vtm}=\mathbb{E}_{(v,t)\in\mathcal{B}}\left[H(\mathbf{p}_{vtm}(v,t),\mathbf{y}_{*}(v,t))\right]
\end{equation}
$\mathbf{y}_*$ denotes soft-label in terms of \textit{intra}, \textit{inter} or \textit{joint}. Finally, fuse $L_{vtm}$ by:
\begin{equation}
    \mathcal{L}_{final-vtm} = (1-\beta) \cdot L_{vtm} + \beta \cdot L_{soft-vtm}
\end{equation}
where $\beta$ is hyperparameter balancing vanilla vtm and soft vtm.

\subsection{Training Objective}

Combining video-text alignment and video-text matching, we formulate the training objective as following:
\begin{equation}
    \mathcal{L} = \mathcal{L}_{vta} + \mathcal{L}_{final-vtm}.
\end{equation}

\section{Experiments}
\subsection{Datasets and Evaluation Metric}
We report our experimental results on five public available video-text retrieval datasets. Following \cite{luo2021clip4clip}, we use recall at rank $K$ (R@$K$), median rank (MdR) and mean rank (MnR) as metrics to evaluate our model.

\noindent\textbf{MSR-VTT} \cite{xu2016msrvtt} is the most popular benchmark for video-text retrieval task. It consists of 10000 videos with 20 captions for each. We report our results on the standard full split \cite{dzabraev2021mdmmt} and 1K-A split \cite{yu2018joint}. The former contains 6513 videos for training, 497 videos for validating and 2990 videos for testing. And the latter uses 9000 videos for training and 1000 video-text pairs for testing.

\noindent\textbf{MSVD} \cite{chen2011collecting} collects 1970 videos with 1200 for training, 100 for validation and 670 for testing. Each video has a length of during 1 to 62 seconds and is paired with approximately 40 sentences.

\noindent\textbf{VATEX} \cite{wang2019vatex} is a large-scale multilingual video description dataset including over 41,250 videos and 825,000 captions. For fair comparison, we follow evaluation protocal used in HGR \cite{chen2020fine} with 25,991 videos for training, 1500 videos for validation and another 1500 videos for testing.

\noindent\textbf{DiDeMo} \cite{anne2017localizing} includes 10,611 videos collected from Flicker and length of each is a maximum of 30 seconds. Following \cite{liu2019use}, \cite{lei2021less}, \cite{bain2021frozen}, we treat it as a video-paragraph retrieval task that all of the captions belonging to the same video will be concatenated to form a paragraph.

\noindent\textbf{ActivityNet} \cite{krishna2017dense} is comprised of 19,994 videos downloaded from YouTube. We follow \cite{zhang2018cross}, \cite{gabeur2020multi} to concatenate all of the video related descriptions to form a paragraph for retrieval and evaluate our model on 'val1' split.

\subsection{Implementations Details}
The basic video encoder and text encoder are both initialized by CLIP \cite{radford2021clip}. Besides basic encoders, frame position embedding is initialized with position embedding used in CLIP's text encoder. Both temporal transformer and multi-modal transformer are initialized by the part of layers from text encoder in CLIP. Concretely, the width, heads, and layers are $512$, $8$ and $4$, respectively. The fixed video length is $12$ and caption length is $32$ for MSR-VTT \cite{xu2016msrvtt}, MSVD \cite{chen2011collecting}, VATEX \cite{wang2019vatex}. As for DiDeMo \cite{anne2017localizing} and ActivityNet \cite{krishna2017dense}, we set both video length and caption length to $64$. $\beta$ and $\tau_2$ for similarity distillation are $0.5$ and $0.5$, respectively. Following \cite{luo2021clip4clip}, we finetune our model with Adam optimizer in $10$ epochs with $256$ batch size and $8$ for $K$. The learning rate is $1e-7$ for the basic video encoder and text encoder, and $1e-4$ for new layers such as frame position embedding, temporal transformer and multi-modal encoder.

\subsection{Comparison with State-of-the-Art Methods}

\begin{table}[ht]
	\begin{center}
    	\caption{Retrieval result on MSR-VTT 1kA split. \textbf{SD} means Similarity Distillation. And the CLIP2TV-ViT16 is always combined with SD. We copy the results of other methods from corresponding papers. Recall at rank $1$(R@1)$\uparrow$, rank $5$(R@5)$\uparrow$, rank $10$(R@10)$\uparrow$, Median Rank(MdR)$\downarrow$ and Mean Rank(MnR)$\downarrow$ are reported.}
    	\label{tab:MSRVTT-1k}
		\resizebox{0.98\textwidth}{!}{
		\begin{tabular}{c|c|ccccc|ccccc} 
	    \hline
        \multicolumn{1}{c}{} & \multicolumn{1}{c}{} & \multicolumn{5}{c}{Text-to-Video} &\multicolumn{5}{c}{Video-to-Text} \\
        \hline
	    Type & Method & R@1 & R@5 & R@10 & MdR & MnR & R@1 & R@5 & R@10 & MdR & MnR \\
        \hline
	    \multicolumn{1}{c|}{\multirow{8}{*}{\rotatebox[origin=c]{90}{Others}}} & JSFusion \cite{yu2018joint}             & 10.2 & 31.2 & 43.2 & 13.0 & -    & -    &    - & -    &   - & - \\
	    & HT-pretrained \cite{miech2019howto100m} & 14.9 & 40.2 & 52.8 & 9.0  & -    & -    &    - & -    &   - & - \\
	    & CE \cite{liu2019use}                    & 20.9 & 48.8 & 62.4 & 6.0  & 28.2 & 20.6 & 50.3 & 64.0 & 5.3 & 25.1 \\
	    & ClipBERT \cite{lei2021less}             & 22.0 & 46.8 & 59.9 & 6.0 & - & - & - & - & - & - \\
	    & TACo \cite{yang2021taco}                & 26.7 & 54.5 & 68.2 & 4.0  & -    & -    & -    & -    & -   & - \\
	    & MMT-pretrained \cite{gabeur2020multi}   & 26.6 & 57.1 & 69.6 & 4.0  & 24.0 & 27.0 & 57.5 & 69.7 & 3.7 & 21.3 \\
	    & SUPPORT-SET \cite{patrick2020support}   & 27.4 & 56.3 & 67.7 & 3.0  & -    & 26.6 & 55.1 & 67.5 & 3.0 & - \\
	    & FROZEN \cite{bain2021frozen}            & 31.0 & 59.5 & 70.5 & 3.0  & -    & -    &    - & -    &   - & - \\
	    & HIT-pretrained \cite{liu2021hit}        & 30.7 & 60.9 & 73.2 & 2.6  & -    & 32.1 & 62.7 & 74.1 & 3.0 & - \\
	    \hline
	    \hline
	    \multicolumn{1}{c|}{\multirow{6}{*}{\rotatebox[origin=c]{90}{CLIP-based}}} & CLIP-straight \cite{portillo2021straightforward} & 31.2 & 53.7 & 64.2 & 4.0  & -    & 27.2 & 51.7 & 62.6 & 5.0 & - \\
	    & MDMMT \cite{dzabraev2021mdmmt}          & 38.9 & 69.0 & 79.7	& 2.0  & 16.5 & -    &    - & -    &   - & - \\
	    & CLIP4Clip-meanP \cite{luo2021clip4clip} & 43.1 & 70.4 & 80.8	& 2.0  & 16.2 & 43.1 & 70.5 & 81.2 & 2.0 & 12.4 \\	
        & CLIP4Clip-seqTransf \cite{luo2021clip4clip} & 44.5 & 71.4 & 81.6	& 2.0  & 15.3 & 42.7 & 70.9 & 80.6 & 2.0 & 11.6 \\
	    & CLIP2Video \cite{fang2021clip2video} & 45.6 & \textbf{72.6} & 81.7 & 2.0 & \textbf{14.6} & 43.5 & 72.3 & 82.1 & 2.0 & \textbf{10.2} \\
	    \hdashline & 70.8 & 81.1 & 2.0 & 15.4 & 45.6 & 72.3 & 83.2 & 2.0 & 11.5 \\ \cdashline{2-12}
	   \multicolumn{1}{c|}{\multirow{3}{*}{\rotatebox[origin=c]{90}{Ours}}} & \textbf{CLIP2TV} & 45.6 & 71.1 & 80.8 & \textbf{2.0} & 15.0 & \textbf{43.9} & 70.9 & \textbf{82.2} & \textbf{2.0} & 12.0 \\
	   & \textbf{CLIP2TV+SD} & \textbf{46.1} & 72.5 & \textbf{82.9} & \textbf{2.0} & 15.2 & 43.9 & \textbf{73.0} & 82.8 & \textbf{2.0} & 11.1 \\
	   \cdashline{2-12}
	   & \textbf{CLIP2TV-ViT16} & \textbf{49.3} & \textbf{74.7} & \textbf{83.6} & \textbf{2.0} & \textbf{13.5} & \textbf{46.9} & \textbf{75.0} & \textbf{85.1} & \textbf{2.0} & \textbf{10.0} \\
    	\hline
		\end{tabular}}
	\end{center}
\end{table}

\begin{table}[t]
    \begin{minipage}{0.48\linewidth}
    \begin{center}
    \caption{MSR-VTT full}
    \label{tab:msrvtt full}
    \resizebox{0.98\textwidth}{!}{
    \begin{tabular}{c|ccccc}
    \hline
    \multicolumn{1}{c|}{\multirow{2}{*}{Method}} & \multicolumn{5}{c}{Text-to-Video} \\ \cline{2-6}
     & R@1 & R@5 & R@10 & MdR & MnR \\
    \hline
    CE \cite{liu2019use} & 10.0 & 29.0 & 41.2 & 16.0 & 86.2 \\
    HT-pretrained \cite{miech2019howto100m} & 14.9 & 40.2 & 52.8 & 9.0 & - \\
    CLIP-straight \cite{portillo2021straightforward} & 21.4 & 41.1 & 50.4 & 10.0 & - \\
    MDMMT \cite{dzabraev2021mdmmt} & 23.1 & 49.8 & 61.8 & 6.0 & 52.8 \\
    CLIP2Video \cite{fang2021clip2video} & 29.8 & 55.5 & 66.2 & 4.0 & 45.5 \\
    \hline
    \textbf{CLIP2TV} & 29.6 & 55.4 & 66.0 & 4.0 & 48.2 \\
    \textbf{CLIP2TV+SD} & \textbf{29.9} & \textbf{55.6} & \textbf{66.3} & \textbf{4.0} & 48.1 \\
    \hdashline
    \textbf{CLIP2TV-ViT16} & \textbf{32.4} & \textbf{58.2} & \textbf{68.6} & \textbf{3.0} & \textbf{43.6} \\
    \hline
    \end{tabular}}
    \end{center}
    \end{minipage}
    \begin{minipage}{0.48\linewidth}
    \begin{center}
    \caption{MSVD}
    \label{tab:msvd}
    \resizebox{0.98\textwidth}{!}{
    \begin{tabular}{c|ccccc}
    \hline
    \multicolumn{1}{c|}{\multirow{2}{*}{Method}} & \multicolumn{5}{c}{Text-to-Video} \\ \cline{2-6}
     & R@1 & R@5 & R@10 & MdR & MnR \\
    \hline
    CE \cite{liu2019use} & 19.8 & 49.0 & 63.8 & 6.0 & - \\
    SUPPORT-SET \cite{patrick2020support} & 28.4 & 60.0 & 72.9 & 4.0 & - \\
    Frozen \cite{bain2021frozen} & 33.7 & 64.7 & 76.3 & 3.0 & - \\
    CLIP-straight \cite{portillo2021straightforward} & 37.0 & 64.1 & 73.8 & 3.0 & - \\
    CLIP4Clip-meanP \cite{luo2021clip4clip} & 46.2 & 76.1 & 84.6 & 2.0 & 10.0 \\
    CLIP4Clip-seqTransf \cite{luo2021clip4clip} & 45.2 & 75.5 & 84.3 & 2.0 & 10.3 \\
    CLIP2Video \cite{fang2021clip2video} & \textbf{47.0} & \textbf{76.8} & \textbf{85.9} & 2.0 & \textbf{9.6} \\
    \hline
    \textbf{CLIP2TV} & 46.3 & 76.1 & 85.3 & 2.0 & 10.0 \\
    \textbf{CLIP2TV+SD} & \textbf{47.0} & 76.5 & 85.1 & 2.0 & 10.1 \\
    \hdashline
    \textbf{CLIP2TV-ViT16} & \textbf{50.2} & \textbf{79.8} & \textbf{87.9} & \textbf{1.0} & \textbf{8.6} \\
    \hline
    \end{tabular}}
    \end{center}
    \end{minipage}
\end{table}

\begin{table}[t]
    \begin{minipage}{0.48\linewidth}
    \begin{center}
    \caption{DiDeMo}
    \label{tab:didemo}
    \resizebox{0.98\textwidth}{!}{
    \begin{tabular}{c|ccccc}
    \hline
    \multicolumn{1}{c|}{\multirow{2}{*}{Method}} & \multicolumn{5}{c}{Text-to-Video} \\ \cline{2-6}
     & R@1 & R@5 & R@10 & MdR & MnR \\
    \hline
    CE \cite{liu2019use} & 16.1 & 41.1 & - & 8.3 & 43.7 \\
    ClipBERT \cite{lei2021less} & 20.4 & 48.0 & 60.8 & 6.0 & - \\
    Frozen \cite{bain2021frozen} & 34.6 & 65.0 & 74.7 & 3.0 & - \\
    CLIP4Clip-meanP \cite{luo2021clip4clip} & 43.4 & 70.2 & 80.6 & 2.0 & 17.5 \\
    CLIP4Clip-seqTransf \cite{luo2021clip4clip} & 43.4 & 69.9 & 80.2 & 2.0 & 17.5 \\
    CLIP4Clip-meanP* \cite{luo2021clip4clip} & 42.1 & 69.3 & 79.6 & 2.0 & 18.0 \\
    \hline
    \textbf{CLIP2TV} & 43.9 & \textbf{70.5} & 79.8 & 2.0 & \textbf{16.6} \\
    \textbf{CLIP2TV+SD} & \textbf{45.5} & 69.7 & \textbf{80.6} & \textbf{2.0} & 17.1 \\
    \hline
    \end{tabular}}
    \end{center}
    \end{minipage}
    \begin{minipage}{0.48\linewidth}
    \begin{center}
    \caption{ActivityNet}
    \label{tab:activitynet}
    \resizebox{0.98\textwidth}{!}{
    \begin{tabular}{c|ccccc}
    \hline
    \multicolumn{1}{c|}{\multirow{2}{*}{Method}} & \multicolumn{5}{c}{Text-to-Video} \\ \cline{2-6}
     & R@1 & R@5 & R@50 & MdR & MnR \\
    \hline
    CE \cite{liu2019use} & 18.2 & 47.7 & 91.4 & 6.0 & 23.1 \\
    ClipBERT \cite{lei2021less} & 21.3 & 49.0 & - & 6.0 & - \\
    SUPPORT-SET \cite{patrick2020support} & 26.8 & 58.1 & 93.5 & 3.0 & -\\
    MMT-pretrained \cite{gabeur2020multi} & 28.7 & 61.4 & 94.5 & 3.3 & 16.0 \\
    HIT-pretrained \cite{liu2021hit} & 29.6 & 60.7 & 95.6 & 3.0 & - \\
    CLIP4Clip-meanP \cite{luo2021clip4clip} & 40.5 & 72.4 & 98.1 & 2.0 & 7.4 \\
    CLIP4Clip-seqTransf \cite{luo2021clip4clip} & 40.5 & 72.4 & 98.2 & 2.0 & 7.5 \\
    CLIP4Clip-meanP* \cite{luo2021clip4clip} & 40.0 & 71.2 & 98.0 & 2.0 & 7.7 \\
    \hline
    \textbf{CLIP2TV} & 40.8 & 72.9 & 98.0 & 2.0 & 7.3 \\
    \textbf{CLIP2TV+SD} & \textbf{44.1} & \textbf{75.2} & \textbf{98.4} & \textbf{2.0} & \textbf{6.5} \\
    \hline
    \end{tabular}}
    \end{center}
    \end{minipage}
\end{table}

\begin{table}[t]
    \begin{center}
    \caption{VATEX. CLIP4Clip-seqTransf* means there are no results about VATEX of HGR split reported in \cite{luo2021clip4clip} and \cite{fang2021clip2video}, so we conduct it by ourselves.}
    \label{tab:vatex}
    \footnotesize
    \setlength{\tabcolsep}{6pt}{
    \begin{tabular}{c|ccccc}
    \hline
    Method & R@1 & R@5 & R@10 & MdR & MnR \\
    \hline
    CLIP-straight \cite{portillo2021straightforward} & 39.7 & 72.3 & 82.2 & 2.0 & 12.8 \\
    SUPPORT-SET \cite{patrick2020support} & 44.9 & 82.1 & 89.7 & 1.0 & - \\
    CLIP4Clip-seqTransf* \cite{luo2021clip4clip} & 60.3 & 89.1 & 94.4 & 1.0 & 4.5 \\
    CLIP2Video \cite{fang2021clip2video} & 61.2 & 90.9 & 95.6 & 1.0 & \textbf{3.4} \\
    \hline
    \textbf{CLIP2TV} & 61.4 & 90.6 & 95.2 & 1.0 & 3.7 \\
    \textbf{CLIP2TV+SD} & \textbf{61.5} & \textbf{90.9} & \textbf{95.6} & 1.0 & 3.7 \\
    \hdashline
    \textbf{CLIP2TV-ViT16} & \textbf{65.4} & \textbf{92.7} & \textbf{96.6} & 1.0 & \textbf{2.9} \\
    \hline
    \end{tabular}}
    \end{center}
\end{table}

We compare our model with other state-of-the-art methods. Table.\ref{tab:MSRVTT-1k} shows the text-to-video and video-to-text retrieval results of CLIP2TV on MSR-VTT 1kA.
As non-CLIP-based methods are usually trained beyond feature-level, which means models adopt video frame features extracted by off-the-shelf pretrained visual models as inputs, they cannot exceed the zero-shot performance of CLIP which is trained end-to-end with much more data.
We categorize these approaches into CLIP-based and others for a fair comparison.
On MSR-VTT 1kA, we achieve state-of-the-art results on both text-to-video and video-to-text retrieval.
Upon strong baseline of CLIP4clip, CLIP2TV improves R1 of text-to-video to 45.6, and R1 of video-to-text to 43.9.
CLIP2TV+SD shows the results with similarity distillation. It further improves CLIP2TV on nearly on all metrics.
Particularly, when replace ViT-32 with ViT-16, we can acquire higher performance(with 3.2\% improvement in text-to-video retrieval at R@1).

Table.\ref{tab:msrvtt full}-Table.\ref{tab:vatex} present the text-to-video retrieval results on MSR-VTT full, MSVD, DiDeMo, ActivityNet and VATEX. For DiDeMo and ActivityNet, we reproduce results based on the provided code of CLIP4Clip and denote as CLIP4Clip-meanP*. On MSR-VTT full, MSVD, and VATEX, we obtain comparable performance to previous SOTA CLIP2Video with ViT32.
On MSVD, though the temporal transformer hardly learns temporal representation with very few frames \cite{luo2021clip4clip,fang2021clip2video}, we still boost the performance close to CLIP2Video.
We also note that on VATEX, similarity distillation boosts the performance with small margin. This is mainly due to the reason that captions in VATEX are relatively detailed and distinct, which just validates our claim.
Our method outperforms previous SOTA methods by a large margin on paragraph-to-video retrieval on DiDeMo and ActivityNet, which further demonstrates our superiority on longer videos with longer descriptions.
Since our method is orthogonal to CLIP2Video, it is expected that combining CLIP2TV with CLIP2Video might further improve the results. We leave this for future study.

\subsection{Ablation Study}

\begin{table}[ht]
    \begin{center}
    \caption{Evaluation of Multi-Modal Fusion with vanilla vtm on different datasets. Recall at rank $1$ (R@1)$\uparrow$ and Mean Rank (MnR)$\downarrow$ of text-to-video retrieval are reported. M-V and Anet mean MSR-VTT and ActivityNet, respectively. The underline indicates that retrieval results are inferred from this module.}
    \label{tab:ablation on vtm}
    \setlength{\tabcolsep}{2.5pt}{
    \begin{tabular}{cc|cc|cc|cc|cc|cc|cc}
        \hline
        \multicolumn{1}{c}{\multirow{2}{*}{vta}} & \multicolumn{1}{c|}{\multirow{2}{*}{vtm}} & \multicolumn{2}{c|}{M-V 1kA} & \multicolumn{2}{c|}{M-V full} & \multicolumn{2}{c|}{MSVD} & \multicolumn{2}{c|}{VATEX} & \multicolumn{2}{c|}{DiDeMo} & \multicolumn{2}{c}{Anet} \\
         & & R@1 & MnR & R@1 & MnR & R@1 & MnR  & R@1 & MnR & R@1 & MnR & R@1 & MnR \\
         \hline\hline
        \checkmark &   & 43.8 & 15.8 & 29.2 & 49.5 & 45.8 & 11.1 & 60.3 & 4.5 & 41.6 & 18.9 & \textbf{42.2} & \textbf{7.4} \\
         & \checkmark & 30.4 & 67.4 & 17.3 & 357.4 & 31.7 & 16.1 & 56.2 & 4.0 & 16.7 & 53.2 & 0.2 & 1851.2 \\
        \uline{\checkmark} & \checkmark & \textbf{45.6} & \textbf{15.0} & \textbf{29.6} & \textbf{48.2} & \textbf{46.3} & \textbf{10.0} & \textbf{61.4} & \textbf{3.7} & \textbf{43.9} & \textbf{16.6} & 40.8 & \textbf{7.3} \\
        \checkmark & \uline{\checkmark} & 39.9 & - & 26.2 & - & 37.2 & - & 59.0 & - & 34.6 & - & 33.5 & - \\
        \hline
    \end{tabular}}
    \end{center}
\end{table}

\textbf{Why alignment with matching?} To evaluate the effectiveness of alignment and matching, we test various combinations of vta and vtm. The ablation results are presented in Table.\ref{tab:ablation on vtm}. 
The 1st row shows the result of vta without vtm. It can be seen as fine-tuning CLIP on other datasets. The result is still competitive due to the outstanding ability of CLIP pretrained on huge data.

However, as shown in the 2nd row of Table.\ref{tab:ablation on vtm}, when we remove the alignment part and train our model with only vtm, the result is severely degraded. Worse on ActivityNet, the training is totally decayed. We attribute this catastrophe to the nature of the dataset. On the one hand, videos in ActivityNet are usually longer than 60 seconds, and some videos can be as long as 100 seconds or even more than 200 seconds, so that input frames with fixed length of 64 cannot cover all the scenes described in paragraph. On the other hand, the paragraph paired with a video consists of several sentences only describing the corresponding video segments, while such video segments may be discontinuous in the video, which means some input frames sampled from the interval of two segments are invalid. These two aspects make it difficult for multi-modal encoder to model the relationship between text words and video frames. This justifies the necessity of aligning two modalities before sending them to multi-modal transformer.

The 3rd row is CLIP2TV combining vta with vtm, retrieving the result from vta. The result shows that it consistently outperforms vta except on ActivityNet.
The last row shares the same structure with the 3rd row, but retrieves the result from vtm. For inference efficiency, we only select top-200 candidates from vta as inputs to vtm for re-ranking. Since the ground truth target might not be in the 200 candidates, the mean ranking of ground truth will be absent from vtm in such case, therefore we neglect MnR results here. We can see that inference result from vtm is much worse than it from vta. This observation is different from \cite{li2021align} on image-text retrieval. We speculate the reason is vtm pretrained on images and texts lacks enough videos for further pretrain.

To sum up, vta and vtm work in a coordinated manner and benefit each other. The alignment is essential and vta can generate hard negatives to vtm. In reverse, vtm tunes vta more finely via back propagating gradients. Moreover, since we retrieve the targets using vta, CLIP2TV avoids computing candidates in vtm with the heavy load, and is practically friendly.

\noindent\textbf{What is the best negative sampling strategy for vtm?} As we know, negative samples plays a crucial role in contrastive learning. Will a larger number of negatives bring benefits as in self-supervised learning? With these questions, we design thorough experiments to find the answer. We choose MSR-VTT 1kA with noisy captions and VATEX with more detailed descriptions. We use $K$ to denote number of negatives. Results shown in Table.\ref{tab:ablation on K} demonstrate that with the increase of $K$, it brings more noise and ambiguity in pairs which hurts the performance. The model decays or saturates when the number of negative samples reaches a certain amount. As MSR-VTT has more noisy samples, a small $K$ is good for it, while VATEX can support larger $K$ value.

\begin{table}[ht]
    \begin{center}
    \caption{Evaluation of the number of negative pairs $K$. We report our experiment results on MSR-VTT 1kA and VATEX.}
    \label{tab:ablation on K}
    \setlength{\tabcolsep}{3.0pt}{
    \begin{tabular}{c|ccccc|ccccc}
        \hline
        \multicolumn{1}{c|}{\multirow{2}{*}{K}} & \multicolumn{5}{c|}{MSR-VTT 1kA} & \multicolumn{5}{c}{VATEX} \\
         & R@1 & R@5 & R@10 & MdR & MnR & R@1 & R@5 & R@10 & MdR & MnR \\
         \hline\hline
        4 & 45.1 & 71.1 & 81.7 & 2.0 & 15.3 & 60.7 & 90.4 & 95.4 & 1.0 & 3.7 \\
        \hline
        8 & 45.6 & 71.1 & 80.8 & 2.0 & 15.0 & 61.4 & 90.6 & 95.2 & 1.0 & 3.7 \\
        \hline
        16 & 44.5 & 71.0 & 81.1 & 2.0 & 14.5 & 61.3 & 90.6 & 95.4 & 1.0 & 3.8 \\
        \hline
    \end{tabular}}
    \end{center}
    \vspace{-6mm}
\end{table}

\begin{figure}
    \centering
    \includegraphics[width=0.98\linewidth]{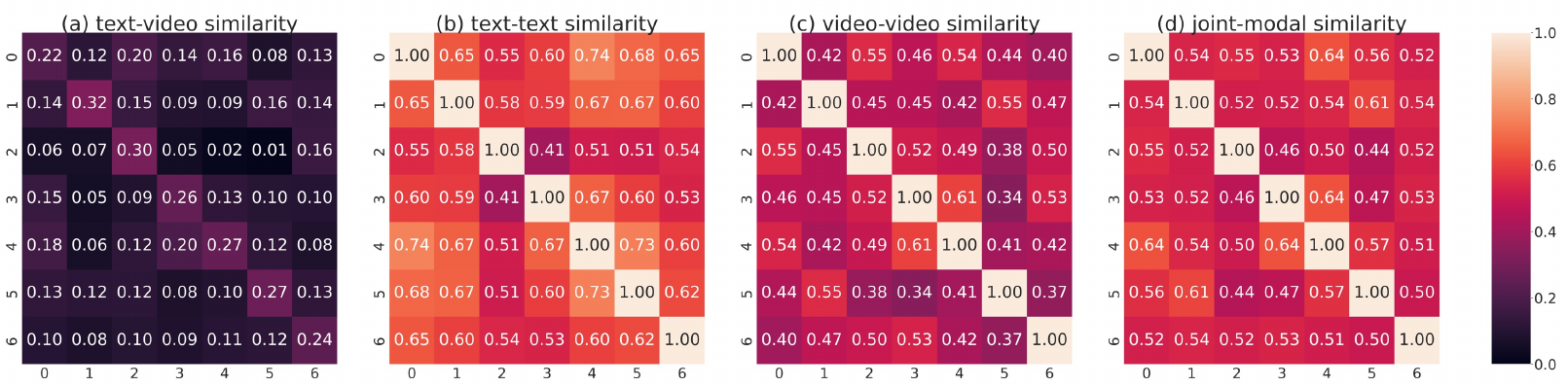}
    \caption{Visualized similarity matrix. (a) text-video similarity. (b) text-text similarity. (c) video-video similarity. (d) joint-modal similarity which is average of text-text and video-video similarity.}
    \label{fig:heatmap}
\end{figure}

\noindent\textbf{How to distill similarity for vtm?} 
We show performance of three variants of similarity distillation on MSR-VTT 1kA in Table.\ref{tab:ablation on soft label}. 
We can see that all three variants can improve the results above vanilla vtm.
Specifically, results from joint and intra are nearly the same, and intra is slightly better than inter. This implies that intra-modal similarity reflects more accurate relative distances between an anchor representation of text/video, towards a group of video/text representations. Moreover, joint inner-modal metric from video and text is more accurate than single modal. To better illustrate the difference of intra- and inter-modal similarity, we show similarity matrices in Figure.\ref{fig:heatmap}, we can see that overall similarity values are gradually increasing along (a)text-video, (b)video-video, and (c)text-text. (d) is the average similarity of text-text and video-video. 

In Table.\ref{tab:ablation on vtm type}, we show the results of vanilla vtm, soft-vtm and mean of them by changing the value of $\beta$, across datasets of MSR-VTT, DiDeMo and ActivityNet. Since these datasets contain relatively more data noise, thus similarity distillation contributes more on them. Also, combining vanilla vtm and soft-vtm achieves the best results. We leave more choices of $\beta$ for future study.

\begin{table}[ht]
    \begin{center}
    \caption{Evaluation of soft-label variants. We report our experimental results on the MSR-VTT 1kA.}
    \label{tab:ablation on soft label}
    \setlength{\tabcolsep}{10.0pt}{
    \begin{tabular}{c|ccccc}
        \hline
        soft-label & R@1 & R@5 & R@10 & MdR & MnR \\
        \hline
        inter & 45.8 & 72.5 & 81.6 & 2.0 & 14.6 \\
        \hline
        intra & 45.9 & 72.3 & 81.5 & 2.0 & 15.3 \\
        \hline
        joint & 46.1 & 72.5 & 82.9 & 2.0 & 15.2 \\
        \hline
    \end{tabular}}
    \end{center}
\end{table}

\begin{table}[ht]
    \begin{center}
    \caption{Evaluation of soft vtm. We report our experimental results on MSR-VTT 1kA, DiDeMo and ActivityNet.}
    \label{tab:ablation on vtm type}
    \resizebox{\textwidth}{!}{
    \begin{tabular}{c|cccc|cccc|cccc}
        \hline
        \multicolumn{1}{c|}{\multirow{2}{*}{vtm}} & \multicolumn{4}{c|}{MSR-VTT 1kA} & \multicolumn{4}{c|}{DiDeMo} & \multicolumn{4}{c}{ActivityNet} \\
         & R@1 & R@5 & R@10 & MnR & R@1 & R@5 & R@10 & MnR & R@1 & R@5 & R@50 & MnR \\
        \hline\hline
        vanilla & 45.6 & 71.1 & 80.8 & 15.0 & 43.9 & 70.5 &79.8 & 16.6 & 40.8 & 72.9 & 98.0 & 7.3 \\
        \hline
        soft($\beta=1.0$) & 44.8 & 71.8 & 82.3 & 15.2 & 42.7 & 68.3 & 79.3 & 17.5 & 43.4 & 74.7 & 98.1 & 6.8 \\
        \hline
        soft($\beta=0.5$) & 46.1 & 72.5 & 82.9 & 15.2 & 45.5 & 69.7 & 80.6 & 17.1 & 44.1 & 75.2 & 98.4 & 6.5 \\
        \hline
    \end{tabular}}
    \end{center}
\end{table}

\subsection{Qualitative Results}

\begin{figure}[ht]
    \centering
    \includegraphics[width=0.98\linewidth]{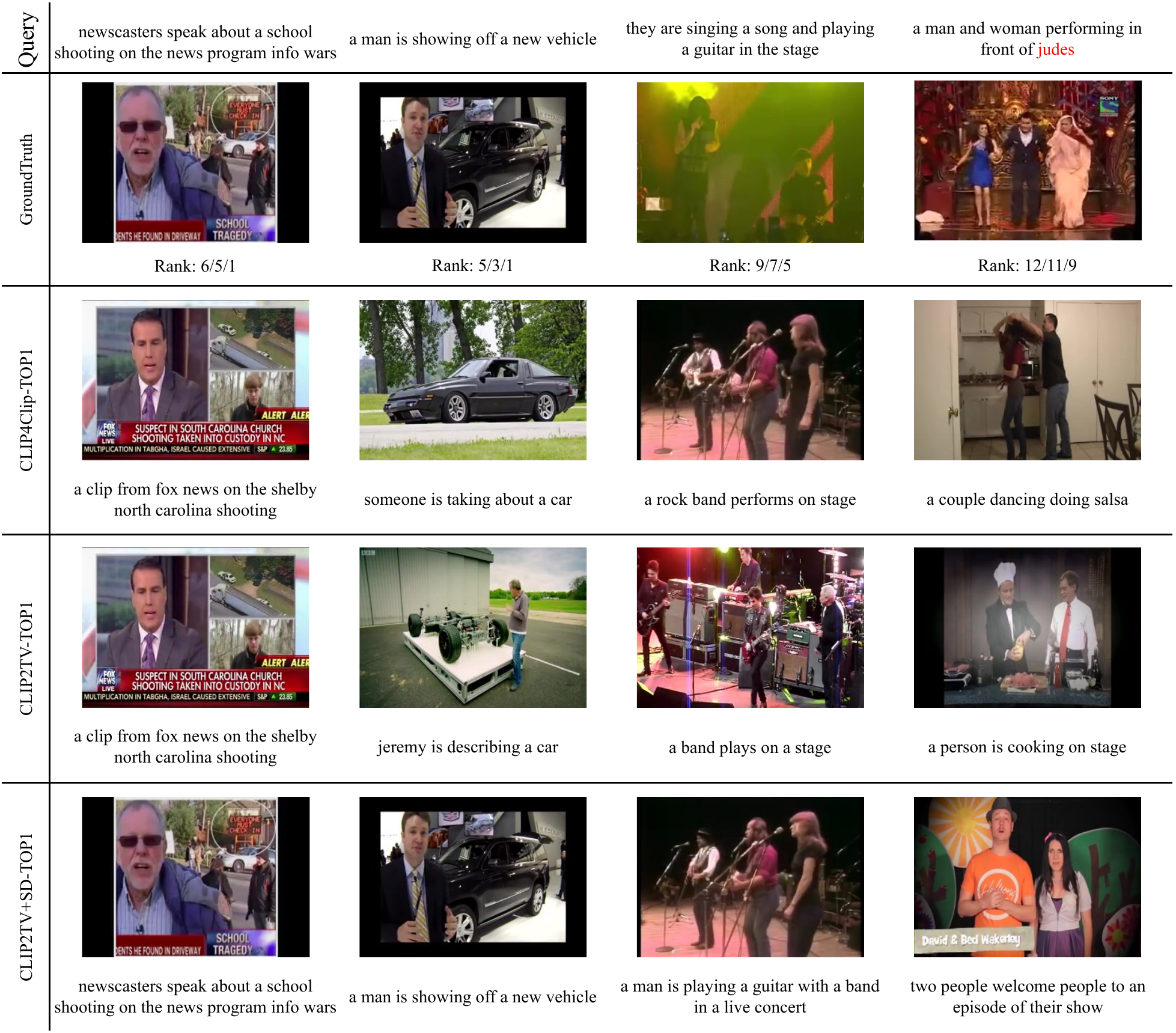}
    \caption{Visualized text-to-video retrieval results on MSR-VTT 1kA. The 1st row is the query, and the 2nd row is the corresponding ground-truth video followed by ranks acquired from CLIP4Clip, CLIP2TV, and CLIP2TV+SD, respectively. The 3rd-5th rows are the retrieved top-1 video by each method and its paired caption. The word ``judes'' in red is a typo.}
    \label{fig:qualitative}
\end{figure}

To illustrate the advantage of the proposed method and the impact of noise, we visualize some text-to-video retrieval results on MSR-VTT 1kA shown in Fig.\ref{fig:qualitative}. The 1st row is the text for query, and the 2nd row is the corresponding ground-truth video followed by its retrieval rank obtained from CLIP4Clip, CLIP2TV, and CLIP2TV+SD, respectively. The 3rd-5th rows present the retrieved R@1 video by each method and its paired caption. 

As shown in Fig.~\ref{fig:qualitative}, our proposed CLIP2TV can perform better and obtain more accurate retrieval results than CLIP4Clip. With the help of similarity distillation from joint soft-label on vtm module, CLIP2TV+SD can make a great improvement when querying a detailed text. On the other hand, CLIP2TV+SD is more robust to vague (3rd column of Fig.\ref{fig:qualitative}) and typo (4th column of Fig.\ref{fig:qualitative}) query, which illustrates that  similarity distillation can alleviate the negative effects of data noise.
\section{Conclusions}
In this work, we propose CLIP2TV, a new CLIP-based framework on video-text retrieval.
CLIP2TV is composed of a video-text alignment module and a video-text matching module.
The two modules are trained in a coordinated manner and benefit each other. To address the problem brought by data noise from popular datasets, we propose similarity distillation and explore different variants based in intra- and inter-modal space. To validate our motivation and explore best practice in this field, we conduct extensive experiments across several datasets. Experimental results prove the effectiveness of ours and we achieve better or competitive results towards previous SOTA methods. CLIP2TV is fast in inference and orthogonal to current CLIP-based frameworks, thus can be easily combined with them. We believe our work can bring insights and practical expertise to both community and industry.



\clearpage

%
%
\bibliographystyle{splncs04}
\bibliography{egbib}

\end{document}